\documentclass[lettersize,journal]{IEEEtran}
\usepackage{amsmath,amsfonts}
\usepackage{algorithmic}
\usepackage{algorithm}
\usepackage{array}
\usepackage[caption=false,font=normalsize,labelfont=sf,textfont=sf]{subfig}
\usepackage{textcomp}
\usepackage{stfloats}
\usepackage{url}
\usepackage{verbatim}
\usepackage{graphicx}
\usepackage{multirow}
\usepackage{cite}
\usepackage{xcolor}
\usepackage{mathtools}

\usepackage{cases}
\usepackage[scr=boondoxo,scrscaled=1.05]{mathalfa}
\usepackage[capitalise]{cleveref}

\begin{document}

\title{  Minimizing Worst-Case Violations of \\ Neural Networks}

\author{Rahul~Nellikkath,~\IEEEmembership{Student~Member,~IEEE,}
        Spyros~Chatzivasileiadis,~\IEEEmembership{Senior~Member,~IEEE}
\thanks{\noindent This work is supported by the FLEXGRID project, funded by the European Commission Horizon 2020 program, Grant Agreement No. 863876, and by the ERC Starting Grant VeriPhIED, Grant Agreement No. 949899}
\thanks{R. Nellikkath and S. Chatzivasileiadis are with the Department of Wind and Energy Systems, Technical University of Denmark (DTU), Kgs. Lyngby, Denmark. E-mail: \{rnelli, spchatz\}@dtu.dk.} 
}



\maketitle

\begin{abstract}
Machine learning (ML) algorithms are remarkably good at approximating complex non-linear relationships. Most ML training processes, however, are designed to deliver ML tools with good \emph{average} performance, but do not offer any guarantees about their worst-case estimation error. For safety-critical systems such as power systems, this places a major barrier for their adoption.
So far, approaches could determine the worst-case violations of only \emph{trained} ML algorithms. To the best of our knowledge, this is the first paper to introduce a neural network training procedure designed to achieve both a good average performance and minimum worst-case violations.
Using the Optimal Power Flow (OPF) problem as a guiding application, our approach (i) introduces a framework that reduces the worst-case generation constraint violations \emph{during training}, incorporating them as a differentiable optimization layer; and (ii) presents a neural network sequential learning architecture to significantly accelerate it. We demonstrate the proposed architecture on four different test systems ranging from 39 buses to 162 buses, for both AC-OPF and DC-OPF applications.

\end{abstract}


\section{Introduction}
\IEEEPARstart{O}{ptimal} Power Flow (OPF) is a valuable tool used by power system operators, electricity market operators, and others actors to evaluate multiple operational or investment scenarios, determine electricity market bidding strategies, or find optimal control setpoints. However, in its original form, the AC- OPF formulation is non-linear and non-convex \cite{CARPENTIER1979, CARPENTIER1985}, making the exact AC-OPF problem formulation sometimes intractable and often time-consuming to  solve \cite{ChallengesOPF}. This encouraged researchers to develop multiple convex approximations and relaxations of the AC-OPF problem \cite{ConvexRelax, ConvexRelax2, LPOPF}. The most popular are the linearized DC-OPF approximations, which ensure fast computation speed and convergence guarantees at the cost of reduced accuracy (especially in highly loaded systems) \cite{DCOPF}.

In that respect, machine learning algorithms, and especially neural networks, have appeared as a promising new tool, as they have been shown to achieve a speedup of \mbox{100-1000 times} compared with the traditional solution methods for the \mbox{AC-OPF} \cite{ReviewMLSec, SurveyML, recent}. This implies that in the time it would take to evaluate a single scenario by solving an AC-OPF instance, we could assess up to 1000 scenarios. Besides that, trained ML models have also shown considerable promise as fast surrogate functions in place of intractable constraints or in bi-level optimization problems to make them computationally feasible \cite{Closing}. These developments have led researchers to focus on the development of advanced ML architectures, and especially neural networks (NN), with improved prediction accuracy for power system applications. One of the promising developments among them is, for example, the Physics-Informed Neural Networks (PINNs) which incorporate the physical equations governing the power flow into NN training \cite{Sensitivity, PINN1, PINN2, PINN3, PINN4, PINN5}. PINNs can achieve higher prediction accuracy while using fewer training data points. 

However, the main challenge remains: ML algorithms act as a black box, and it is hard to assess their worst-case performance across the whole set of possible inputs once deployed. Considering that AC-OPF could be used for safety-critical applications, this poses a major barrier for the widespread adoption of ML algorithms in power systems. 
Our previous work \cite{Andreas} proposed a method to quantify the worst-case constraint violation caused by a \emph{trained} NN across the entire input domain. We extended this work by developing a procedure to choose the hyperparameters that minimized the worst-case generation constraint violations of a PINN for DC-OPF \cite{MYDC} and AC-OPF problems \cite{MYAC}. However, to the best of our knowledge, so far, there has been no neural network training procedure that incorporates the minimization of the worst-case violations \emph{across the entire input domain} inside the training; such a procedure would be able to deliver NNs that by-design achieve both an accurate average performance and reduced worst-case violations.

Traditionally, the issue of constraint violation was tackled by restraining the neural network output to the system constraints. For instance, a method was proposed in \cite{Constraint1} and \cite{Constraint2} where the generation setpoints predicted by the NN were post-processed to avoid generation constraint violation. However, this affects the accuracy of the NN prediction. Alternatively, generation constraint violations can be penalized during NN training through an additional term in the NN training loss function to minimize the average generation constraint violation \cite{Constraint3, Constraint1}. However, since the loss function is usually formulated to reduce the \emph{average} error in the NN training set, such methods usually minimize the average generation constraint violation but do not directly impact the worst-case constraint violations. 

To overcome this challenge, a NN training architecture that prioritizes the worse performing data points in the training set by using the conditional value at risk (CVaR) loss function was proposed in \cite{CVaRisk}. 
Similarly, a Min-Max algorithm to minimize the maximum loss in the training set of a standard NN (i.e. estimation error) was proposed in\cite{MinMax}. However, these algorithms rely heavily on the quality and quantity of the training dataset. The training dataset will have to cover well both probable and improbable operating points for those algorithms to be effective. Such datasets usually do not exist, and are often challenging to generate. Furthermore, although they do reduce the worst-case violations, none of these proposed methods can provide any worst-case performance guarantees across the entire continuous input domain. 

This paper proposes a NN training architecture that integrates an explicit procedure to minimize the worst-case violations \emph{during} training and across the entire input domain. Our contributions are the following:
\begin{enumerate}
    \item We introduce a NN training architecture that incorporates the NN worst-case constraint violations as a differentiable layer in NN. Using the OPF as a guiding application and focusing on the generator constraints, our NN training architecture is able to train the NN not only minimizing the average prediction error as usual, but also, at the same time, minimizing the worst-case violations of the generator constraints \emph{during} training. 
    \item The proposed architecture captures the entire input domain and does not rely on the training dataset to minimize worst-case constraint violations; it is, therefore, not dependent on the quantity or quality of the training data. It is also modular and can be easily integrated with most of the existing NN training algorithms that use the ReLU activation function.
    \item To reduce computation time, we transform the proposed training algorithm to a NN sequential learning architecture, which is able to reduce the worst-case violations of a trained NN without impacting its generalization capabilities.
    \item We demonstrate the effectiveness of the proposed NN learning architecture for both DC and AC optimal power flow problems, on four different test systems from PGLib-OPF network library v19.05, ranging from 39 buses to 162 buses. 
\end{enumerate}
The remainder of this paper is structured as follows: Section II describes the DC-OPF and AC-OPF problem, and the use of neural networks for OPF. Section III introduces the algorithm used to determine the worst-case guarantees, and the proposed neural network training architectures. Section~\ref{sec:results} presents the simulation setup and the results demonstrating the performance of the proposed neural network training architecture. Section IV concludes. 

\section{Optimal Power Flow and Neural Networks}
This section discusses the fundamentals for the AC-OPF and DC-OPF problems, which will be used as a guiding application for the Neural Network Training Architectures proposed in this paper, present the basic neural network architecture, and introduces the necessary terminology.
\subsection{Optimal Power Flow Problem}
\subsubsection{AC-Optimal Power Flow}
The objective function that minimizes the cost of active and reactive power generation in a power system  with $N_g$ number of generators, $N_b$ number of buses, and $N_d$ number of loads can be formulated as follows:

\begin{equation}
    \min_{\mathbf{P}_g,\mathbf{Q}_g, \mathbf{v}} \quad \mathbf{c}^T_p\mathbf{P}_g
    \label{obj}
\end{equation}
where vector $\mathbf{P}_g$ denotes the active power setpoints of every generator in the system, and $\mathbf{c}^T_p$ collects the costs for the active power production of each generator. $\mathbf{Q}_g, \mathbf{v}$ are optimization variables and denote the reactive power setpoints of every generator and the complex bus voltages, respectively. The active and reactive power injection at each node $n \in N_b$ can be represented as follows:
\begin{align}
    p_n = p_n^g - &p_n^d  &\forall n\in N_b, \label{P_n} \\
    q_n = q_n^g - &q_n^d  &\forall n\in N_b, \label{Q_n}
\end{align}
where $p_n$ and $q_n$ denote the active and reactive power injection at node $n$, ${p}_n^g$ and ${p}_n^d$ indicate the active power generation and demand at node $n$, and ${q}_n^g$ and ${q}_n^d$ denote the reactive power generation and demand, respectively at node $n$. The power flow equations in the network can be expressed as follows:
\begin{align}    
    p_n = \sum_{k=1}^{N_b} v_n^r(v_k^r G_{nk} - v_k^i B_{nk}) + v_n^i(v_k^i G_{nk} + v_k^r B_{nk}) \\
    q_n = \sum_{k=1}^{N_b} v_n^i(v_k^r G_{nk} - v_k^i B_{nk}) - v_n^r(v_k^i G_{nk} + v_k^r B_{nk})
\end{align}
where $v^r_n$ and $v^i_n$ denote the real and imaginary part of the voltage at node $n$. The conductance and susceptance of the line $nk$ connecting node $n$ and $k$ is denoted by $G_{nk}$ and $B_{nk}$ respectively. The power flow equations can be simplified by combining the real and imaginary parts of voltage into the vector $\mathbf v = [(\mathbf v^r)^T , (\mathbf v^i)^T]^T$, of size $2N_b \times 1$, as follows \cite{Sensitivity}:
\begin{align}
    \mathbf{v}^T \mathbf{M}_{p}^n \mathbf{v} = &p_n        & \forall n\in N_b \label{PF_p}\\
    \mathbf{v}^T \mathbf{M}_{q}^n \mathbf{v} = &q_n            & \forall n\in N_b \label{PF_q}
\end{align}
where $\mathbf{M}_{p}^n$ and $\mathbf{M}_{q}^n$ are symmetric real valued matrices \cite{formulation}. Then the active and reactive power generation limits can be formulated as follows:
\begin{align}
    \underline{p}_n^g \leq &\mathbf{v}^T \mathbf{M}_{p}^n \mathbf{v} + p_n^d \leq \overline{p}_n^g & \forall n\in N_g  \label{Pg_lim} \\
    \underline{q}_n^g \leq &\mathbf{v}^T \mathbf{M}_{q}^n \mathbf{v} + q_n^d\leq \overline{q}_n^g & \forall n\in N_g \label{Qg_lim}
\end{align}

Similarly, the voltage and line current flow constraints for the power system can be represented as follows: 
\begin{align}
    \underline{\mathbf V}^n \leq \mathbf{v}^T \mathbf{M}_{v}^n \mathbf{v} &\leq \overline{\mathbf V}^n & \forall n\in N_b \label{V_lim} \\
    {\ell}_{mn} = \mathbf{v}^T \mathbf{M}_{i_{mn}} & \mathbf{v} \leq \overline{\mathbf \ell}_{mn} & \forall mn\in N_l \label{I_lim}
\end{align}
where $\mathbf{M}_{v}^n := e_ne^T_n + e_{N_b+n}e_{N_b+n}^T$ and $e_n$ is a $2N_b \times 1$ unit vector with zeros at all the locations except $n$. The squared magnitude of upper and lower voltage limit is denoted by $\overline{\mathbf V}^n$ and $\underline{\mathbf V}^n$ respectively. The squared magnitude of line current in line $mn$ is represented by ${\ell}_{mn}$ and matrix  $\mathbf{M}_{i}^{mn} = |y_{mn}|^2(e_m - e_n)(e_m -e_n)^T + |y_{mn}|^2(e_{N_b+m} - e_{N_b+n})(e_{N_b+m} - e_{N_b+n})^T$, where $y_{mn}$ is the line admittance of branch $mn$. Assuming the slack bus $N_{sb}$ acts as the angle reference for the voltage, we will have:
\begin{equation}
    v^i_{N_{sb}} = \mathbf{v}^T \mathbf{e}_{N_b + N_{sb}} \mathbf{e}^T_{N_b +N_{sb}} \mathbf{v} = 0 
    \label{V_sb}
\end{equation}

The constraints \eqref{P_n}-\eqref{Q_n},\eqref{PF_p}-\eqref{V_sb} and the objective function for the AC-OPF problem \eqref{obj} can be simplified as follows (for more details, see \cite{Sensitivity}): 
\begin{subequations}
\begin{align}
    \min_{\mathbf{v},\mathbf G} & \quad \mathbf{c}^T \mathbf G \label{pri_1}\\
    s.t \text{ } \mathbf{v}^T \mathbf{L}_l \mathbf{v} &= a_l^T \mathbf G +b_l^T \mathbf{D}, & l=1:L \label{pri_2}\\
    \mathbf{v}^T \mathbf{M}_m \mathbf{v} &\leq d_m^T\mathbf{D} + f_m, & m = 1 :M \label{pri_3}
\end{align}
\label{pri_full}
\end{subequations}
where $\mathbf{G} = [\mathbf{P}_g^T , \mathbf{Q}_g^T]^T$, and $\mathbf{c}^T$ is the combined linear cost terms for the active power and, if necessary, reactive power generation. $\mathbf{D} = [\mathbf{P}_d^T , \mathbf{Q}_d^T]^T$ denotes the active and reactive power demand in the system. Following that, the equality constraints \eqref{PF_p}-\eqref{PF_q} and \eqref{V_sb} can be represented by the $\mathbf L=2N_b+1$ constraints in \eqref{pri_2}. Similarly, the inequality constraints \eqref{Pg_lim}-\eqref{I_lim} can be represented by the $\mathbf M=4N_g+2N_b+N_l$ constraints in \eqref{pri_3}.

In its original form, the AC-OPF formulation given in \eqref{pri_full}, is a non-convex Quadratically Constrained (QC) problem. To speed up the computation time and ensure convergence, the linearized "DC Optimal Power Flow" (DC-OPF) is frequently used. This offers reliably a quick estimate, although often not too accurate. We detail DC-OPF in the next paragraph. 
\subsubsection{DC - Optimal Power Flow}
DC-OPF is a linear approximation of the AC-OPF problem. A DC-OPF problem for generation cost minimization in a system with $N_{b}$ number of buses, $N_g$ number of generators, and $N_{d}$ number of loads can be represented as follows:
\begin{equation}
    \min_ {\mathbf{P}_{g}} \quad \mathbf{c}^T_p  \mathbf{P}_{g}
    \label{Obj}
\end{equation}
\begin{equation}
    \sum_{i=1}^{N_g} p_n^g - \sum_{i=1}^{N_d} p_n^d = 0
\label{ConLoadBal}
\end{equation}
\begin{equation}
    \mathbf{\underline{P}}_{g} \leq \mathbf{P}_{g} \leq \mathbf{\overline{P}}_{g}
    \label{ConGenLim}
\end{equation}
\begin{equation}
    \mathbf{\underline{P}}_{d} \leq \mathbf{P}_{d} \leq \mathbf{\overline{P}}_{d}
    \label{ConDenLim}
\end{equation}
\begin{equation}
\lvert \mathbf{PTDF}(\mathbf{P}_{g} - \mathbf{P}_{d}) \lvert \leq \mathbf{\overline{P}}_{l}
\label{ConLineLim}
\end{equation}
where the maximum and minimum active power generation limits are denoted by $\mathbf{\overline{P}}_g$ and $\mathbf{\underline{P}}_g$, respectively. $\mathbf{\overline{P}}_d$ and $\mathbf{\underline{P}}_d$ denote the maximum and minimum active power demand. $\mathbf{PTDF}$ is the power transfer distribution factors (for more details, see \cite{LecturenoteOPF}), and $\mathbf{\overline{P}}_{l}$ denotes the active power flow line limit.

The objective function for minimizing active power generation cost is given by \eqref{Obj}. Constraint \eqref{ConLoadBal} ensures the balance between active power supply and demand. The active power generation, demand, and line flow limits are given in \eqref{ConGenLim} - \eqref{ConLineLim}, respectively, with \eqref{ConLineLim} representing the linearized power flows.

\subsection{Neural Networks for Optimal Power Flow Approximation}\label{SecPINN}
Neural Networks (NNs) are considered global approximators. Assuming they have sufficient size and are trained appropriately, they can determine the AC-OPF and DC-OPF solution without loss of accuracy. Neural Networks are comprised of a set of interconnected hidden layers with multiple neurons that learn the relationship between the input and output (in the case of our OPF, the input is the power system demand, and the output is the optimal generation setpoints). Each neuron in a hidden layer is connected with neurons in the neighboring layers through a set of edges. The information exiting one neuron goes through a linear transformation over the respective edge before reaching the neuron in the subsequent layer. Inside every neuron, a non-linear so-called ``activation function'' is applied on the information.

\begin{figure}[htbp]
\centerline{\includegraphics[scale=.4]{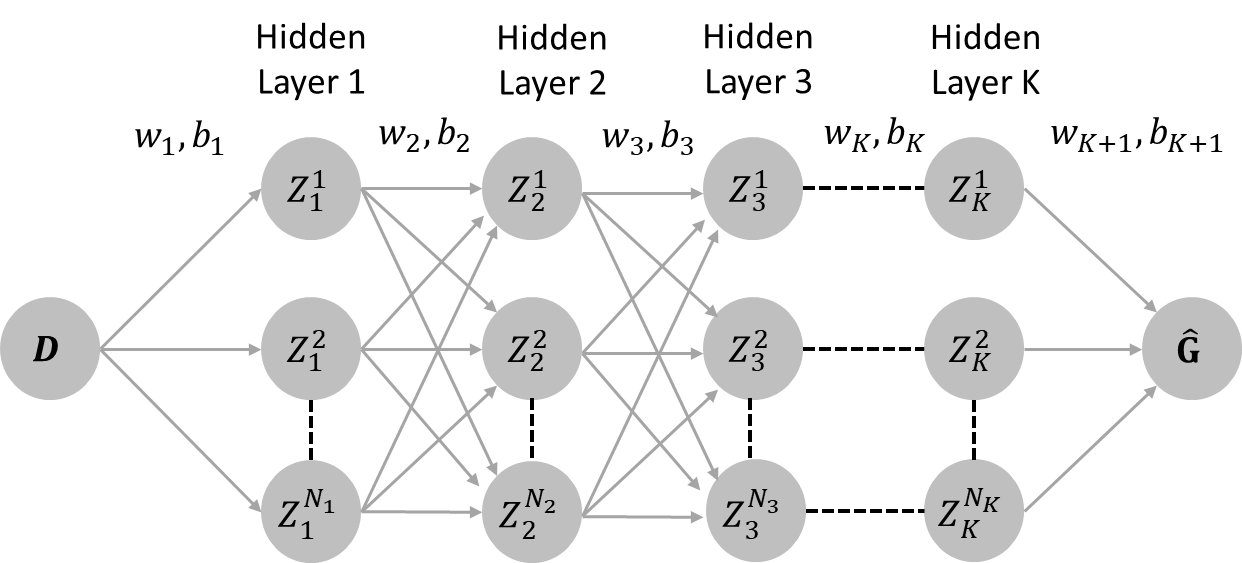}}
\caption{Illustration of the neural network architecture to predict the optimal generation active and reactive power outputs $\mathbf{\hat G}$ using the active and reactive power demand $\mathbf{D}$ as input: There are K hidden layers in the neural network with $N_k$ neurons each. Where k = 1, ...,K.}
\label{NN_basic}
\end{figure}

For a NN, with $K$ number of hidden layers and $N_k$ number of neurons in hidden layer $k$, as shown in \cref{NN_basic}, the information arriving at layer $k$ can be formulated as follows:
\begin{equation}
   \hat{\mathbf{Z}}_k = \mathbf{w_{k}}\mathbf{Z}_{k-1}+\mathbf{b_{k}}\label{NN1}
\end{equation}
where ${\mathbf{Z}}_{k-1}$ is the output of the neurons in layer $k-1$, $\hat{\mathbf{Z}}_k$ is the information received at layer $k$, $\mathbf{w_{k}}$ and $\mathbf{b_{k}}$ are the weights and biases connecting layer $k-1$ and $k$.

Each neuron applies a nonlinear activation function, as shown in \eqref{eq:sigma}. These functions help neural networks accurately approximate the nonlinear relationships between the input and output layer. The output of each layer in the NN can be represented as follows:
\begin{equation}
    \mathbf{Z}_k = \sigma( \hat{\mathbf{Z}}_k ) \label{eq:sigma}
\end{equation}
where $\sigma$ is the nonlinear activation function. There is a range of possible activation functions, such as the sigmoid function, the hyberbolic tangent, the Rectifier Linear Unit (ReLU), and others. In this paper, we use the ReLU as the activation function, similar to the vast majority of recent papers, as this has been shown to accelerate the neural network training \cite{glorot}. The ReLU activation function can be formulated as follows:
\begin{align}
    \mathbf{Z}_k &= \max( \hat{\mathbf{Z}}_k,0)\label{Relu}
\end{align}
Given an active and reactive power demand of a system, this paper trains NNs to determine the optimal active and reactive power setpoints of the AC-OPF problem, as a guiding application. Similarly, considering a DC-OPF, we also train NNs to determine the optimal active power generation setpoints, receiving the active power demand as an input.
The average error in predicting the optimal generation setpoints in the training data set, denoted by $\mathcal{L}_{0}$, is measured:
\begin{equation}
    \mathcal{L}_{0} = \frac{1}{N} \sum_{i=1}^N | \mathbf{G}_i - \hat{\mathbf{G}}_i| \label{L_0}
\end{equation}
where $N$ is the number of data points in the training set, $\mathbf{G}_i$ is the generation setpoint determined by the OPF, and $\hat{\mathbf{G}}_i$ is the predicted NN generation setpoint.

The backpropagation algorithm modifies the weights and biases in every iteration of the NN training to minimize the average prediction error of generation setpoints in the training set ($\mathcal{L}_{0}$). However, minimizing the average error does not guarantee small constraint violations across the entire input domain. To overcome this challenge, in this paper we propose a method that integrates the evaluation of the worst-case constraint violations during training and includes it in the NN loss function, as we describe in the next section.

\section{Neural Network Training to Minimize Worst-Case Constraint Violations} \label{EWCG}
\subsection{Determining the Worst-Case Constraint Violations}
\label{NNtrain_A}
A novel NN training algorithm was proposed in \cite{MinMax} and \cite{CVaRisk} that reduces worst-case violations by assigning higher penalties on 
the worst-performing data points of the training dataset. However, for these algorithms to be effective, high quality training data sets are necessary, which should cover all possible operating scenarios. This is hard in a high-dimensional input space. 

In this paper, instead of relying on the training dataset, we build upon our previous work \cite{Andreas} and \cite{MYAC}, with the goal to minimize the worst-case constraint violations \emph{during training} (instead of ex-post) and across the entire input domain. Contrary to considering both the worst-case line flow and generation constraint violations as in \cite{Andreas} and \cite{MYAC}, in this paper we focus only on the worst-case generation constraint violations caused by the NN predictions, to reduce the computational burden of the NN training process. 
A similar NN training architecture could be developed to incorporate the worst-case line flow and voltage violations of the NN. Future work will focus on combining the different types of constraint violations in a common, computationally efficient procedure.

The optimization problem for determining the maximum generation constraint violations is formulated as follows:
\begin{subequations}
\begin{gather}
    \underset{\mathbf D}{\mathrm{max}} \quad v_g \label{WCeq1} \\
    v_g = \underset{ \mathbf D}{\mathrm{max}}(\mathbf{\hat{G} - \overline{G}}, \mathbf{\underline{G} - \hat{G}},0) \label{WCeq2}\\
    \text{s.t.} \text{ } \eqref{NN1}, \eqref{Relu}
\end{gather} \label{WCeq}
\end{subequations}
where $\mathbf{\overline{G}}$ is the maximum and $\mathbf{\underline{G}}$ is the minimum active and reactive power generation capacity, respectively. The formulation for the ReLU activation function in \eqref{Relu} is nonlinear, and can be reformulated into a set of mixed-integer linear inequality constraints, as shown in \eqref{RelU1}-\eqref{Relu5}. This renders problem \eqref{WCeq} to a Mixed Integer Linear Program (MILP) \cite{Andreas}. 
\begin{subnumcases}
{\mathbf{Z}_k = \max(\hat{\mathbf{Z}}_k,0)\Rightarrow}
\mathbf{Z}_k  \leq \hat{\mathbf{Z}}_k - \underline{\mathbf{Z}}_k  (1-\mathbf{y}_k) \label{RelU1} \\ 
\mathbf{Z}_k  \geq \hat{\mathbf{Z}}_k \label{RelU2}   \\
\mathbf{Z}_k  \leq \overline{\mathbf{Z}}_k \mathbf{y}_k  \label{RelU3}  \\
\mathbf{Z}_k   \geq \mathbf{0}  \label{Relu4}  \\
\mathbf{y}_k \in \{0,1\}^{N_k} \label{Relu5}.
\end{subnumcases}
where $\hat{\mathbf{Z}}_k$  is the input to the ReLU activation function, $\mathbf{Z}_k$ is its output, and $\mathbf{y}_k$ is a binary variable that signifies the state of the ReLU activation function. When $\hat{\mathbf{Z}}_k$ is less than zero and the ReLU is ``inactive'', $\mathbf{y}$ and $\mathbf{Z}_k$ are zero, and similarly when $\hat{\mathbf{Z}}_k$ is greater than zero and the ReLU is ``active'', $\mathbf{y}$ is one and $\mathbf{Z}_k$ will be equal to $\hat{\mathbf{Z}}_k$. $\overline{\mathbf{Z}}_k$ and $\underline{\mathbf{Z}}_k$ are the maximum and minimum values possible for the respective $\hat{\mathbf{Z}}_k$. They should be large enough so that the constraints \eqref{RelU1} and \eqref{RelU3} will not be binding and small enough so that the constraints are not unbounded. 

\subsection{Simultaneous Neural Network Training for Good Average Performance and Minimum Worst-Case Violations}
\label{NNtrain_B}
The objective function for a neural network training algorithm which can simultaneously minimize the average prediction error (denoted by $\mathcal{L}_{0}$) \emph{and} the worst-case generation constraint violation (denoted by $\mathcal{L}_{wc}$) can be formulated as a bilevel optimization problem, as follows:
\begin{align}
    \min_{\mathbf{w}, \mathbf{b}} \quad \Lambda_0 \mathcal{L}_{0} + \Lambda_w \mathcal{L}_{wc},
    \label{WCNN_obj}
\end{align}
where $\Lambda_0$ and $\Lambda_{wc}$ are the weights assigned to the loss functions $\mathcal{L}_{0}$ and $\mathcal{L}_{wc}$, respectively. $\mathcal{L}_{0}$ corresponds to \eqref{L_0}. $\mathcal{L}_{wc}$ represents the loss function for the worst-case generation constraint violation, which is formulated as a nested (lower-level) optimization problem itself:
\begin{subequations}
\begin{gather}
    \quad \mathcal{L}_{wc} = \underset{\mathbf D, \mathbf{y}}{\mathrm{max}} \quad v_g \label{LWC1}\\
    \text{s.t.} \quad \eqref{NN1},\eqref{WCeq2}, \eqref{RelU1} -\eqref{Relu5} \label{LWC2}
\end{gather} \label{LWCall}
\end{subequations}


Problem \eqref{LWCall} defines a Mixed Integer Linear Program (MILP). Here, the worst-case performance of the NN given in \eqref{LWC1} depends solely on the defined input domain $\mathbf{D}$, and the NN weights $\mathbf{w}$ and biases $\mathbf{b}$, which are determined by the upper-level optimization problem in \eqref{WCNN_obj}. Optimization variables $\mathbf{y}$ are binary variables governed by \eqref{RelU1}-\eqref{Relu5} that help determine if each ReLU function is ``active'' or ``inactive''. 
Problem \eqref{WCNN_obj} with $\Lambda_{w}=0$ represents the standard NN training algorithms. Most such algorithms rely on computing the gradient of the loss function with respect to the weights and biases 
in every iteration, e.g. backpropagation. For $\Lambda_{w} \neq 0$, however, differentiating through the lower-level MILP problem to assess the sensitivity of the worst-case violation to the weights and biases is cumbersome to implement. Instead, we assume that the ReLU status associated with the worst-case generation constraint violation does not change for a small perturbation in weights and biases of the NN. This seemingly arbitrary but logical assumption will be confirmed by the good performance of our algorithms in Section~\label{sec:results}. Following that, we design the following procedure. First, we solve the MILP \eqref{LWCall} as a stand-alone problem to determine $\mathbf{y}$, see \eqref{LWC_new1}.
Then, we fix $\mathbf{y}$ and convert \eqref{LWCall} into the Linear Programming (LP) problem \eqref{LWC_new}. 

First step:
\vspace{-10pt}
\begin{subequations}
\begin{gather}
    \quad  \mathbf{y}^*(\mathbf{w},\mathbf{b},\mathbf{D}) = \mathrm{argmax}_\mathbf{y} v_g \label{LWC1_new1}\\
    \text{s.t.} \quad \eqref{NN1},\eqref{WCeq2}, \eqref{RelU1} -\eqref{Relu5} \label{LWC1_new1}
\end{gather} \label{LWC_new1}
\end{subequations}
Second step:
\vspace{-10pt}
\begin{subequations}
\begin{gather}
    \quad \mathcal{L}_{wc} = \underset{\mathbf D}{\mathrm{max}} \quad v_g(\mathbf{y}^*) \label{LWC1_new}\\
    \text{s.t.} \quad \eqref{NN1},\eqref{WCeq2} \label{LWC2_new}
\end{gather} \label{LWC_new}
\end{subequations}

\begin{figure}[htbp]
\centerline{\includegraphics[scale=.345]{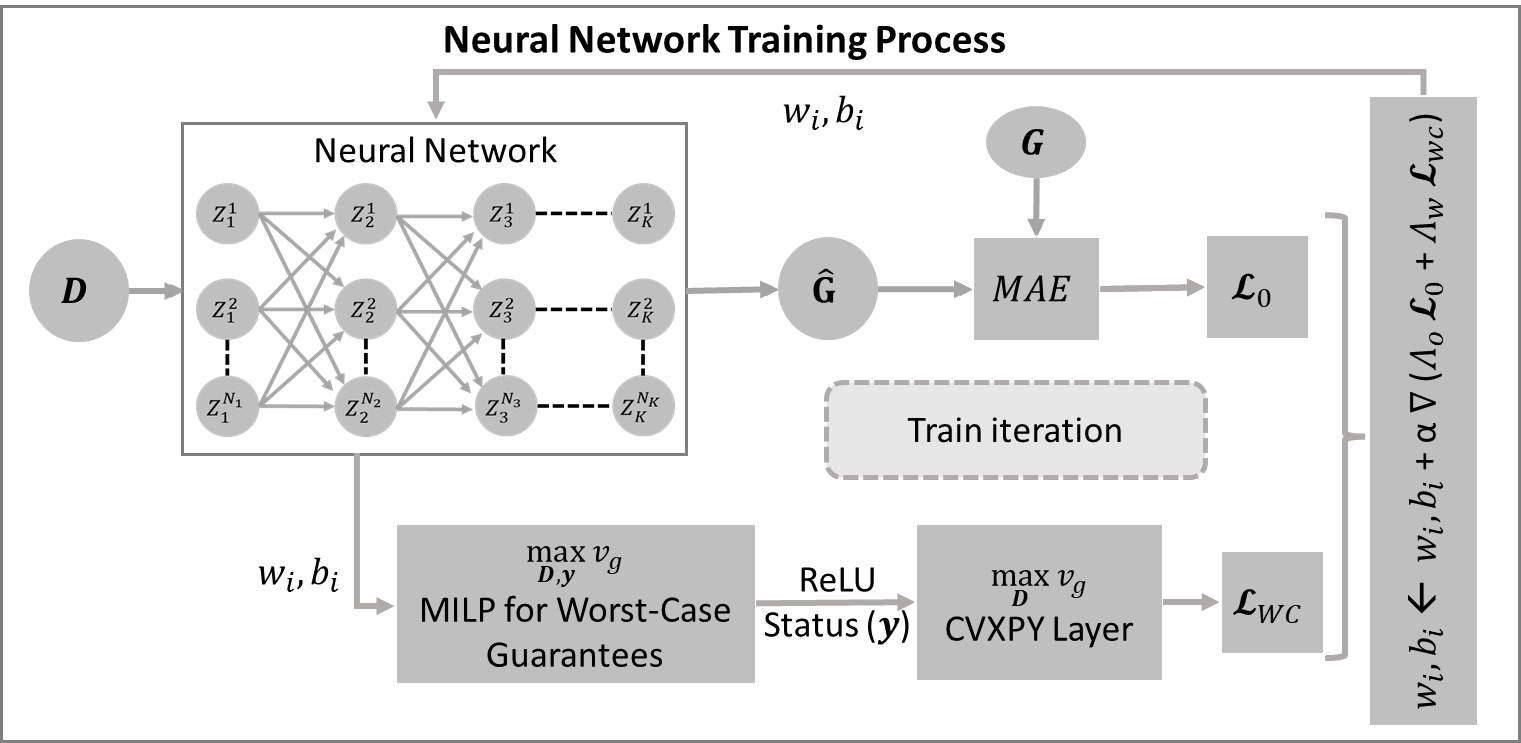}}
\caption{Illustration of the proposed neural network training architecture to minimize worst-case violation of the predicted optimal generation setpoints $\mathbf{\hat G}$ using power demand $\mathbf{D}$ as input. $\mathcal{L}_{0}$ is the average error in predicting the optimal generation setpoints in the training dataset and $\mathcal{L}_{wc}$ is the loss function due to worst-case violation, as denoted in \eqref{LWC_new1}-\eqref{LWC_new}. \cref{alg1} details the proposed neural network training architecture.} 
\label{WCNN}
\end{figure}

The key step in this process is that by converting \eqref{LWCall} to the convex program \eqref{LWC_new}, we can now cast it as a differentiable optimization layer \cite{cvxpylayers}, using the CVXPY toolbox. 
Embedding \eqref{LWC_new} as a differentiable optimization problem allows the NN training optimizer to compute the gradients for both terms of \eqref{WCNN_obj} and backpropagate through them, similar to the standard NN training procedure.
The proposed NN architecture is given in \cref{WCNN} and the detailed algorithm is given in \cref{algo}.
\begin{algorithm}[H]
\caption{Proposed Neural Network Training algorithm}\label{algo}
\begin{algorithmic}[1]
\STATE {\textsc{TRAIN}}$(w_i, b_i)$
\vspace{3pt}
\STATE \hspace{0.5cm} Initialize the NN weights ($w_i$) and biases ($b_i$)
\vspace{3pt}
\STATE \hspace{0.5cm} \parbox[t]{210pt}{ Set the learning rate ($\alpha$), total number of epochs ($T$) and number of iterations before including worst-case violations in the training ($T_{int}$) }
\vspace{3pt}
\STATE \hspace{0.5cm} initialize t=1
\vspace{3pt}
\STATE \hspace{0.5cm} Repeat
\vspace{3pt}
\STATE \hspace{1 cm} \parbox[t]{200pt}{Calculate average loss $\mathcal{L}_0$ in the training dataset}
\vspace{3pt}
\STATE \hspace{1 cm} \parbox[t]{200pt}{Update NN weights and biases $w_i$ and $b_i$ using the back propagation algorithm as follows:\\ $w_i, b_i = w_i,b_i + \alpha \nabla \mathcal{L}_0 $}
\vspace{3pt}
\STATE \hspace{1 cm} t=t+1
\STATE \hspace{0.5cm} until $t = T_{int}$
\vspace{3pt}
\STATE \hspace{0.5cm} Repeat
\vspace{3pt}
\STATE \hspace{1 cm} \parbox[t]{200pt}{Calculate average loss $\mathcal{L}_0$ in the training dataset}
\vspace{3pt}
\STATE \hspace{1 cm} \parbox[t]{200pt}{Solve MILP \eqref{LWC_new1} for worst-case generation constraint violation and obtain the ReLU status}
\vspace{3pt}
\STATE \hspace{1 cm} \parbox[t]{200pt}{Compute $\mathcal{L}_{wc}$ \eqref{LWC_new} using CVXPY Layer}
\vspace{3pt}
\STATE \hspace{1 cm} \parbox[t]{200pt}{Update NN weights and biases $w_i$ and $b_i$ using backpropagation algorithm as follows:\\ $w_i, b_i = w_i,b_i + \alpha \nabla (\Lambda_0 \mathcal{L}_0 +\Lambda_{wc} \mathcal{L}_{wc})$}
\vspace{3pt}
\STATE \hspace{1 cm} t=t+1
\STATE \hspace{0.5cm} until $t = T$
\end{algorithmic}
\label{alg1}
\end{algorithm}

\subsection{Sequential Training to Minimize Worst-case Violations} \label{Sequ}
\begin{figure}[htbp]
\centerline{\includegraphics[scale=.4]{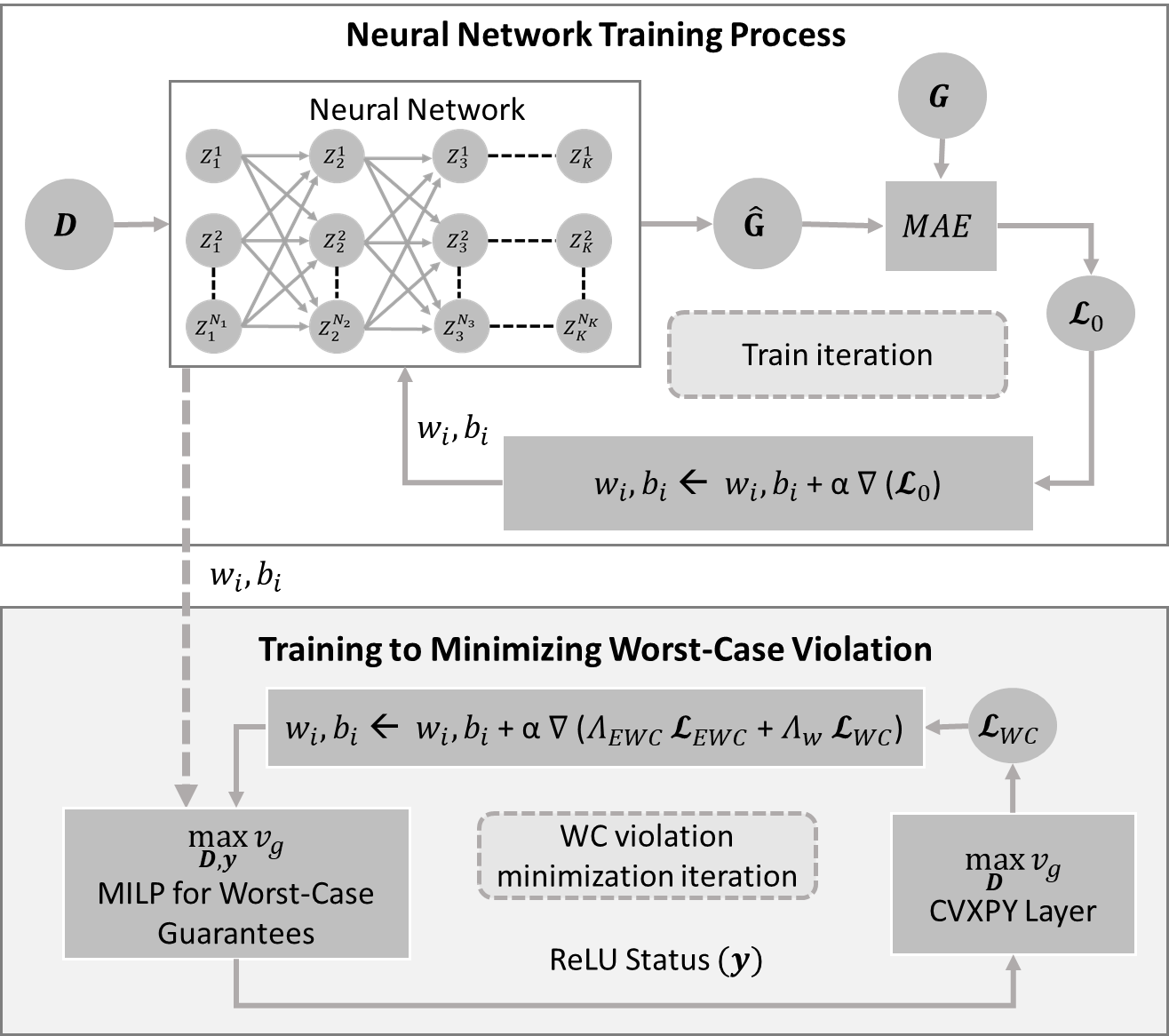}}
\caption{Illustration of the proposed sequential learning neural network training architecture to minimize worst-case violation of the predicted optimal generation set points $\mathbf{\hat G}$ using power demand $\mathbf{D}$ as input. $\mathcal{L}_{0}$ is the average error in predicting the optimal generation setpoints in the training data set,  $\mathcal{L}_{WC}$ is the loss function due to worst-case violation, as denoted in \eqref{LWC_new1}-\eqref{LWC_new}, and $\mathcal{L}_{EWC}$ is the loss function elastic weight consolidations(EWC) (see \cref{AP_EWC}).} 
\label{WCNN2}
\end{figure}

Solving \eqref{WCNN_obj}, which includes steps \eqref{LWC_new1} and \eqref{LWC_new}, is a complex NN training procedure that imposes a significant computation burden. To accelerate computations, this section proposes to decompose \eqref{WCNN_obj} to its two loss terms: $\mathcal{L}_0$ and $\mathcal{L}_{wc}$. Section~\ref{sec:results} will present results for both the simultaneous and the sequential training algorithms. \cref{WCNN2} presents the proposed sequential NN training architecture. Decomposing \eqref{WCNN_obj}, we first determine $\mathbf{w}, \mathbf{b}$ by minimizing $\mathcal{L}_0$, i.e. $ \min_{\mathbf{w}, \mathbf{b}} \mathcal{L}_{0}$. This corresponds to a standard NN training procedure. Subsequently, we plug the obtained $\mathbf{w}, \mathbf{b}$ of the trained NN to \eqref{LWC_new1} to determine the ReLU status $\mathbf{y}$; then we solve \eqref{LWC_new} considering the determined $\mathbf{w}, \mathbf{b}, \mathbf{y}$. Following that, we have to update the weights and biases $\mathbf{w}, \mathbf{b}$ to reduce the worst-case violations. Considering, however, that $\mathbf{w}, \mathbf{b}$ have already been optimized to minimize the average error (i.e. $\mathcal{L}_0$), arbitrarily changing $\mathbf{w}, \mathbf{b}$ could be detrimental for the NN performance. To minimize the impact of the $\mathbf{w}, \mathbf{b}$ update on $\mathcal{L}_0$, we use Elastic Weight Consolidation (EWC), proposed in \cite{EWC}. The EWC formulation is given in \cref{AP_EWC}.

\section{RESULTS \& DISCUSSION}
\label{sec:results}
An increasing portion of the literature proposes variations of NN training algorithms that include power system violations in the loss function to improve the average NN performance. However, these algorithms fail to provide any worst-case performance improvement guarantees. In this section, the worst-case generation constraint violations and the average performance of the proposed NN algorithm in an unseen test dataset are compared against (i) a standard NN, denoted by NN, which does not include any terms for minimizing worst-case violations, and against (ii) a NN with penalties for generation constraints violations included in the NN loss function, denoted by GenNN, as proposed in \cite{MYDC} and \cite{Constraint3}. The loss function formulation for the GenNN architecture is given in \cref{AP_GenNN}. We present results for AC-OPF and DC-OPF problems on four different test systems: the 39-bus, 57-bus, 118-bus, and 162-bus systems from PGLib-OPF network library v19.05 \cite{PGLib}. The test system characteristics are given in \cref{TC}. 

In all instances, the demand at each node is assumed to be able to vary between 60\% to 100\% of their respective nominal loading as follows:
\begin{gather}
    0.6 \text{ }\overline{\mathbf{D}}\leq \mathbf{D} \leq \overline{\mathbf{D}}
\end{gather}
where $\overline{\mathbf{D}}$ denotes the nominal active and reactive power demand. The sum of the maximum loading over all nodes for each system is given in \cref{TC}. 
\begin{table}[h]
\centering
\caption{TEST CASE CHARACTERISTICS}
\begin{tabular}{lllllll}
\hline\hline
\multirow{2}{*}{Test Case} & \multirow{2}{*}{$N_{b}$} & \multirow{2}{*}{$N_d$} & \multirow{2}{*}{$N_g$} & \multirow{2}{*}{$N_l$} & \multicolumn{2}{l}{Max total load} \\ \cline{6-7} 
                           &                          &                        &                        &                        & MW             & MVA             \\ \hline \hline
case39     & 39  & 21  & 10  & 46    & 6254   & 6626                                                    \\ \hline
case57     & 57  & 42  & 4  & 80   & 1251  & 1375                                                  \\ \hline
case118    & 118 & 99  & 19 & 186   & 4242 & 4537                                                      \\ \hline
case162    & 162 & 113 & 12 & 284   & 7239  & 12005                                                    \\ \hline \hline
\end{tabular}
\label{TC}
\end{table}

For the AC-OPF problem, the active and reactive power demand at each node was assumed to vary independently while generating the datasets. Ten thousand sets of random input values were generated using Latin hypercube sampling \cite{hypercube} for all cases, of which 70\% were allocated to the training dataset, 10\% were allocated to the validation set, and the remaining 20\% were allocated to the unseen test dataset. MATPOWER \cite{MATPOWER} was used to solve the AC-OPF and DC-OPF problem and determine the optimal generation set-points that were included in the datasets for the NN training.

A NN with 3 hidden layers and 15 nodes in each layer is used to predict the AC-OPF and DC-OPF solutions. The ML algorithms were implemented using PyTorch \cite{Pytorch} and Adam optimizer \cite{Adam}; a learning rate of 0.001 was used for training. WandB \cite{wandb} was used for monitoring and tuning the hyper parameters. The MILP problem for obtaining the worst-case violations was programmed in Pyomo \cite{pyomo} and solved using the Gurobi solver \cite{gurobi}. Subsequently, the MILP problem was converted to a linear programming problem by identifying the ReLU status, and transformed to a differentiable optimization layer using the CVXPY toolbox \cite{cvxpylayers}. The NNs were trained in a High-Performance Computing (HPC) server with an Intel Xeon E5-2650v4 processor and 256 GB RAM. The code and datasets to reproduce the results are available online \cite{Code_git}.

\subsection{Simultaneous Training and Minimization of Worst-Case Violations of NN}
\subsubsection{Reducing Algorithm Complexity}
Minimizing the worst case violations requires to consider all weights and biases of the NN, as shown in \eqref{WCNN_obj}-\eqref{LWCall}. This imposes a very significant computational burden. To reduce the computational complexity, we explored if we could reduce the optimization variables, i.e. the number of $w,b$, without significantly affecting the NN performance. Before implementing the proposed algorithm, the importance of weights and biases connecting different layers on the worst-case generation constraint violations are studied for DC-OPF and AC-OPF predictions. We did this to explore the possibility of limiting the optimization variables of our worst-case violation algorithm without significantly impacting the effectiveness. During testing, it was observed that the weights and biases connecting the last hidden layer with the output layer had the largest impact on the worst-case violations compared with all other weights and biases. 

\begin{figure}[htbp]
  \centerline{\includegraphics[scale=.65]{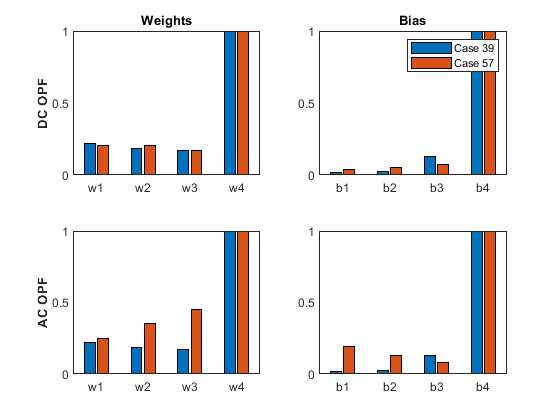}}
  \caption{Normalized mean absolute derivatives of worst-case generation constraint violation w.r.t. the weights (w$i$) and biases (b$i$) connecting different layers.  All values are normalized w.r.t the weights and biases connecting the last hidden layer with the output layer (denoted by w4 and b4) of the NN.}
  \label{Grad}
\end{figure}

 \cref{Grad} presents the mean absolute derivatives that describe the change of the worst generation constraint violation with respect to the weights and biases connecting different layers in a NN, for case39 and case57. In order to remove the starting point bias, the presented results for DC-OPF and AC-OPF problems are the mean values over five different random initializations of NN. 

As we observe, the weights and biases connecting the last hidden layer with the output layer are in all cases considerably more significant than all other weights and biases. As a matter of fact, in most cases the weights and biases between the last layer and the output have an 8 to 10 times higher impact on the worst case violations than between any other layer (i.e. 8$\times$-10$\times$ higher derivative).  This could either be because the weights connecting the last hidden layer to the outer layer are used primarily for scaling the prediction to the output domain in a NN or because the ReLU activation function in the hidden layers makes multiple weights and biases in the hidden layers irrelevant. Future work shall explore this further. Considering this, we limited the optimization variables of our worst-case violation algorithm (see \eqref{WCNN_obj}, \eqref{LWC_new1}, \eqref{LWC_new}) to the weights and biases connecting only the last hidden layer with the output layer. This helps maintain a similarly high efficacy of our algorithm, while substantially reducing its computational complexity and time (reduction of the optimization variables by 75\%). At the same time, this also helped diminish the impact the algorithm has on the NN generalization capability. 
\subsubsection{Minimize Worst-Case Generation Constraint Violations in DC-OPF}
In this section, we train NNs to predict the DC-OPF solutions, given the active power demand as input. We analyze the average and the worst-case performance of the proposed NN algorithm, which we call WCNN, versus a standard NN and a GenNN. The mean absolute error (MAE) over an unseen test data set was used to quantify the average performance of the neural network. MAE in the test set, denoted by $MAE_T$, is evaluated as follows: 
\begin{align}
    MAE_T &= \frac{1}{N_T} \sum_{i=1}^{N_T} \lvert \hat{\mathbf{G}} - \mathbf{G}\lvert \label{eq:MAE}
\end{align}

\begin{figure}[htbp]
  \centerline{\includegraphics[scale=.45]{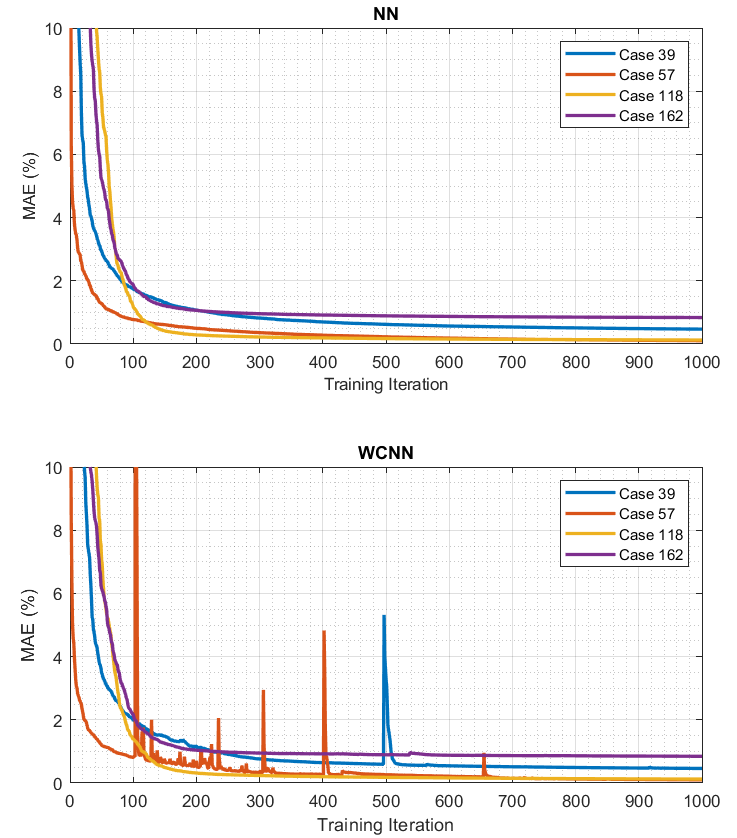}}
  \caption{Mean Absolute Error of the standard NN and the proposed NN (WCNN) in validation set for DC- OPF problem}
  \label{CovRes}
\end{figure}

All three algorithms were trained till convergence, and the MAE of standard NN and WCNN in the validation set after each training iteration is given in \cref{CovRes} (the MAE evolution of the GenNN is similar to the standard NN). Compared to the standard NN, the MAE of the WCNN displayed a few fluctuations in case39 and case57. These fluctuations could be due to the additional loss term for worst-case generation constraint violation. Regardless, these fluctuations plateaued in a few iterations.  The average performance in an unseen test data set and the worst-case performance of all three algorithms after 1000 training iterations are given in \cref{TabDC}.


\begin{table}[h]
\centering
\caption{Average and worst-case performance of the proposed algorithm as compared to the ``standard NN'' and ``NN with generation constraint violations in loss function (GenNN)'' for the DC-OPF problem}
\label{TabDC}
\begin{tabular}{ccccc}
\hline \hline
\multicolumn{2}{c}{\multirow{2}{*}{Test Cases}} & \multirow{2}{*}{MAE (\%)} & \multicolumn{2}{c}{Worst case Guarantees}                                   \\ \cline{4-5} 
\multicolumn{2}{c}{}                            &                           &  $v_g$ (MW) & \begin{tabular}[c]{@{}c@{}}\% wrt\\      max loading\end{tabular} \\ \hline \hline
\multirow{3}{*}{case39}         & NN           & 0.47                                          & 741                         & 11.85                                                                                 \\ \cline{2-5} 
                                 & GenNN        & 0.44                                          & 459                         & 7.34                                                                                  \\ \cline{2-5} 
                                 & WCNN         & 0.46                                          & 419                         & 6.70                                                                                  \\ \hline
\multirow{3}{*}{case57}         & NN           & 0.11                                          & 1708                        & 86.13                                                                                 \\ \cline{2-5} 
                                 & GenNN        & 0.08                                          & 1230                        & 62.03                                                                                 \\ \cline{2-5} 
                                 & WCNN         & 0.10                                          & 361                         & 18.20                                                                                 \\ \hline
\multirow{3}{*}{case118}        & NN           & 0.13                                          & 2374                        & 55.96                                                                                 \\ \cline{2-5} 
                                 & GenNN        & 0.07                                          & 2334                        & 55.02                                                                                 \\ \cline{2-5} 
                                 & WCNN         & 0.12                                          & 1801                        & 42.46                                                                                 \\ \hline
\multirow{3}{*}{case162}        & NN           & 0.84                                          & 12202                       & 168.56                                                                                \\ \cline{2-5} 
                                 & GenNN        & 0.82                                          & 11418                       & 157.73                                                                                \\ \cline{2-5} 
                                 & WCNN         & 0.83                                          & 6985                        & 96.49                                                                                 \\ \hline \hline
\end{tabular}
\end{table}

During testing, we observed that the standard NN accurately predicted the DC-OPF optimal generation set points in all test systems, achieving less than 1\% MAE on the previously unseen test dataset. However, these standard NNs could lead to substantial worst-case generation constraint violations. By including the generation constraint violations in the NN training error (GenNN), the worst-case violations were reduced. However, as the number of generators increases, the GenNNs have a negligible impact on the worst-case generation constraint violations. A similar trend could also be observed in \cite{MYAC}. Perhaps, this is happening because with multiple generators in the system, reducing the generation constraint violation of a single generator becomes much less critical compared with reducing the overall MAE. The underlying reasons that GenNN's effectiveness drops while scaling to larger systems is an object of future work.

In comparison, the proposed NN algorithm (WCNN) resulted in a 30\% to 80\%  reduction of the worst-case generation constraint violations in the test systems. Moreover, the WCNN also positively affected the generalization error in all cases compared with a standard NN. This could be because the WCNN algorithm helped identify a better local optimum for the weights and biases.     
\subsubsection{Minimize Worst-Case Generation Constraint Violations in AC-OPF}
\begin{figure}[htbp]
  \centerline{\includegraphics[scale=.55]{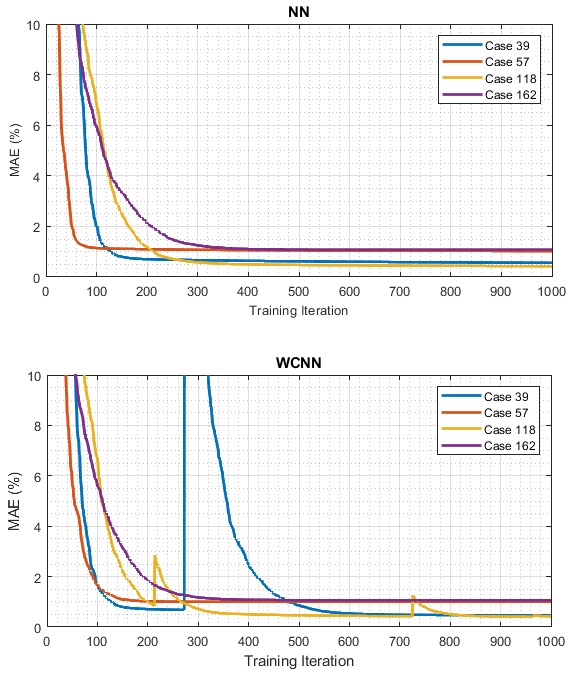}}
  \caption{Mean Absolute Error of the standard NN and the proposed NN (WCNN) in validation set for AC- OPF problem}
  \label{CovRes2}
\end{figure}

This section analyzes the average and the worst-case performance of the proposed WCNN algorithm versus standard NN and GenNN for predicting nonlinear and non-convex AC- OPF solutions. Similar to the previous section, the average prediction error in the unseen test set is assessed using the mean absolute error (MAE), and all three algorithms were trained till convergence. The MAE of the standard NN and the WCNN in the validation set is given in \cref{CovRes2}. 
In case39 the proposed NN algorithm resulted in a significant fluctuation. However, the fluctuation was dampened after a few iterations. All three  algorithms converged after 1000 iterations. The average and worst-case performance of all three algorithms after 1000 training iterations are given in \cref{TabAC}.

\begin{table}[h]
\centering
\caption{Average and worst-case performance of the proposed algorithm as compared to the ``standard NN'' and ``NN with generation constraint violations in loss function (GenNN)'' for the AC-OPF problem}
\label{TabAC}
\begin{tabular}{ccccc}
\hline \hline
\multicolumn{2}{c}{\multirow{2}{*}{Test Cases}} & \multirow{2}{*}{MAE (\%)} & \multicolumn{2}{c}{Worst case Guarantees}                                   \\ \cline{4-5} 
\multicolumn{2}{c}{}                            &                           & $v_g$ (MW) & \begin{tabular}[c]{@{}c@{}}\% wrt\\      max loading\end{tabular} \\ \hline \hline
                                  & NN           & 0.56                                             & 42                          & 0.67                                                                                  \\ \cline{2-5} 
                                  & GenNN        & 0.55                                             & 42                          & 0.67                                                                                  \\ \cline{2-5} 
\multirow{-3}{*}{case39}         & WCNN         & 0.47                                             & 0                           & 0.00                                                                                  \\ \hline
                                  & NN           & 1.02                                             & 13                          & 0.65                                                                                  \\ \cline{2-5} 
                                  & GenNN        & 1.01                                             & 13                          & 0.67                                                                                  \\ \cline{2-5} 
\multirow{-3}{*}{case57}         & WCNN         & 1.00                                             & 6                           & 0.29                                                                                  \\ \hline
                                  & NN           & 0.42                                             & 8679                        & 204.60                                                                                \\ \cline{2-5} 
                                  & GenNN        & 0.42                                             & 9069                        & 213.80                                                                                \\ \cline{2-5} 
\multirow{-3}{*}{case118}        & WCNN         & 0.42                                             & 4659                        & 109.83                                                                                \\ \hline
                                  & NN           & 1.10                                             & 13342                       & 184.30                                                                                \\ \cline{2-5} 
                                  & GenNN        & 1.06 & 13141                       & 181.52                                                                                \\ \cline{2-5} 
\multirow{-3}{*}{case162}        & WCNN         & 1.06                                             & 10305                       & 142.35                                                                                \\ \hline
\end{tabular}
\end{table}
Similar to the DC-OPF case, we observed that the standard NN could make accurate predictions for the AC-OPF optimal generation setpoints in all test systems with less than or equal to 1.1\% MAE in the unseen test dataset. Moreover, the worst-case generation constraint violations in case39 and case57 for the AC-OPF problem were substantially lower than what was observed for the respective DC-OPF predictions. However, in case118 and case162, the worst-case violations of the standard NN (denoted by NN in \ref{TabAC}) are higher than in the DC-OPF problem. Additionally, we observed that GenNNs have a negligible or negative impact on reducing the worst-case generation constraint violations of the trained NN. This further validates our assumption that reducing average generation constraint violation in the training dataset does not guarantee a reduction in the worst-case performance of the NN.

When we compare WCNN with NN or GenNN, the worst-case generation constraint violation was reduced by more than 25\% in all cases, while also having a positive or neutral impact on the MAE. Especially in case39, it is worth highlighting that the proposed WCNN algorithm was able to drive the worst-case generation constraint violation to zero, guaranteeing that the NN is completely safe to operate for all generators across the entire input domain.

To summarize, similar to the DC-OPF case, all results highlight that the proposed NN training architecture to simultaneously minimize the average error and the worst-case violations results in a significant improvement in the worst-case performance of a neural network for the AC-OPF as well. However, this comes at the cost of a significant computation burden. This approach took between 1 to 2 hours to converge compared to the 3 to 5 minutes it would take to train a standard NN. This time is heavily dependent on the number of hidden layers, the number of nodes in each hidden layer, and the number of epochs required for training the NN, and it will grow with increasing NN complexity. An option is to use weight sparsification and ReLU pruning to reduce the NN complexity without affecting its performance \cite{Andreas}. Alternatively, in this paper, we propose to decompose the proposed ``simultaneous'' NN training to a sequential procedure, as detailed in Section \ref{Sequ}. The next Section presents results from its performance.
%

\subsection{Sequential Learning to Minimize Worst-Case Violations}
\begin{figure}[h]
  \centerline{\includegraphics[scale=.45]{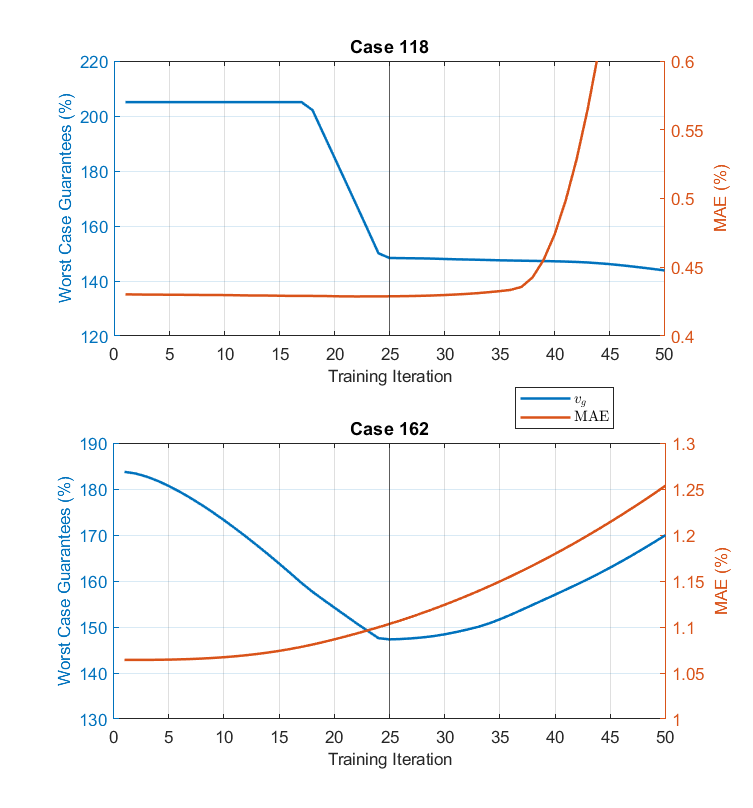}}
  \caption{Worst-case generation constraint violation and mean absolute error in sequential learning to minimize worst-case violations.}
  \label{PPRes}
\end{figure}

%
We implement the sequential learning algorithm, as introduced in \cref{Sequ}, on the trained standard NNs for case118 and case162 and the AC-OPF problem (see Table~\ref{TabAC}). These two NNs were selected because of the massive worst-case generation constraint violations they caused. The weights and biases of the trained NN were passed to the proposed sequential learning block to reduce the worst-case violations. The MAE in the validation set and the worst-case violations, denoted by $v_g$ (and normalized by the maximum demand of the demand scenario they occured), after each iteration are given in \cref{PPRes}.

\cref{PPRes} shows that adjusting the weights and biases of trained NNs to minimize their worst-case violations using the sequential NN training procedure (see the lower panel of \cref{WCNN2}) 
results in a substantial reduction in both cases. 

In case118, the MAE in the test set stayed almost the same during the first 25  iterations, even though we were training to minimize the worst-case violations and did not consider the mean absolute error in our optimization. This probably happens because the proposed sequential NN training algorithm uses Elastic Weight Consolidation, which does consider information from the previously trained task (i.e., MAE minimization). 

In summary, the proposed algorithm achieved a 30\% reduction in worst-case violations in 25 iterations with no significant impact on the MAE in the test set. Training for more iterations resulted in marginal improvement for the worst-case generation constraint violations while increasing the MAE significantly. This could be avoided by using early stopping. 

A similar trend for the MAE and worst-case violations was also observed in case162. However, unlike the previous case, the worst-case violations, instead of stabilizing, increased after 25 iterations. This may have been caused by the Elastic Weight Consolidation (EWC) term (see Appendix~\ref{sec:appendix}) overpowering the worst-case violations. Most importantly, however, as expected, the proposed sequential learning algorithm still resulted in a lower worst-case constraint violation after 25 iterations with a total reduction of more than 25\%.

To summarize, in both tested cases, the proposed sequential learning algorithm achieved a 20\% to 30\% reduction in the worst-case generation constraint violations without significantly impacting the average NN performance. It also appears that ``early stopping'', widely used in optimization and NN training, will be crucial in this procedure to ensure that the validation error does not increase uncontrollably. Equally importantly, this procedure approached the results of the simultaneous learning at a significantly lower computing time, requiring only 10-15 min additional training time.  
\section{Conclusion}
The goal of this paper is to develop Neural Network (NN) training methods designed to deliver both good average performance and worst-case performance guarantees at the same time. To the best of our knowledge, this is the first attempt to develop such methods, both in power systems literature and beyond.  
Neural network training methods that deliver worst-case performance guarantees can build the missing trust and remove barriers for applying Machine Learning methods on safety-critical applications, such as power systems. This paper offers two key contributions. First, we introduce a framework to incorporate the worst-case generation constraint violations of NN algorithms which determine the DC-OPF and AC-OPF optimal solutions. Our NN training framework is designed to minimize the worst-case constraint violation during NN training and deliver at the end a \emph{trained NN with worst-case performance guarantees}. Second, in order to reduce computation time of the proposed framework, we present an NN sequential learning architecture. The proposed method is highly modular and can be used with most existing NN training algorithms during or after training. 
Future research will focus on incorporation of different constraint violations, and on improving computational efficiency through NN weight sparsification and by solutions that could achieve differentiable layers directly for MILP problems or tight convex relaxations of them. 

\appendix \label{sec:appendix}
\subsection{Elastic Weight Consolidation} \label{AP_EWC}
Elastic weight consolidation (EWC) is used to overcome catastrophic forgetting during sequential learning in NN \cite{EWC}. Given a NN which is already trained for a task A and if that same NN has to be trained for a different task B then the loss minimization function for Task B using EWC will be as follows:
\begin{equation}
    \mathcal{L} (\theta) = \mathcal{L}_B (\theta) + \sum_i \Lambda F_i (\theta_i -\theta_{A,i})
\end{equation}
where $\mathcal{L} (\theta)$ is the loss function with EWC, $\mathcal{L}_B (\theta)$ is the loss function for task B, $\Lambda$ is the weight assigned to the EWC loss term, and $F_i$ is the $i^{th}$ diagonal entry of Fisher information matrix. 

\begin{figure}[htbp]
  \centerline{\includegraphics[scale=.35]{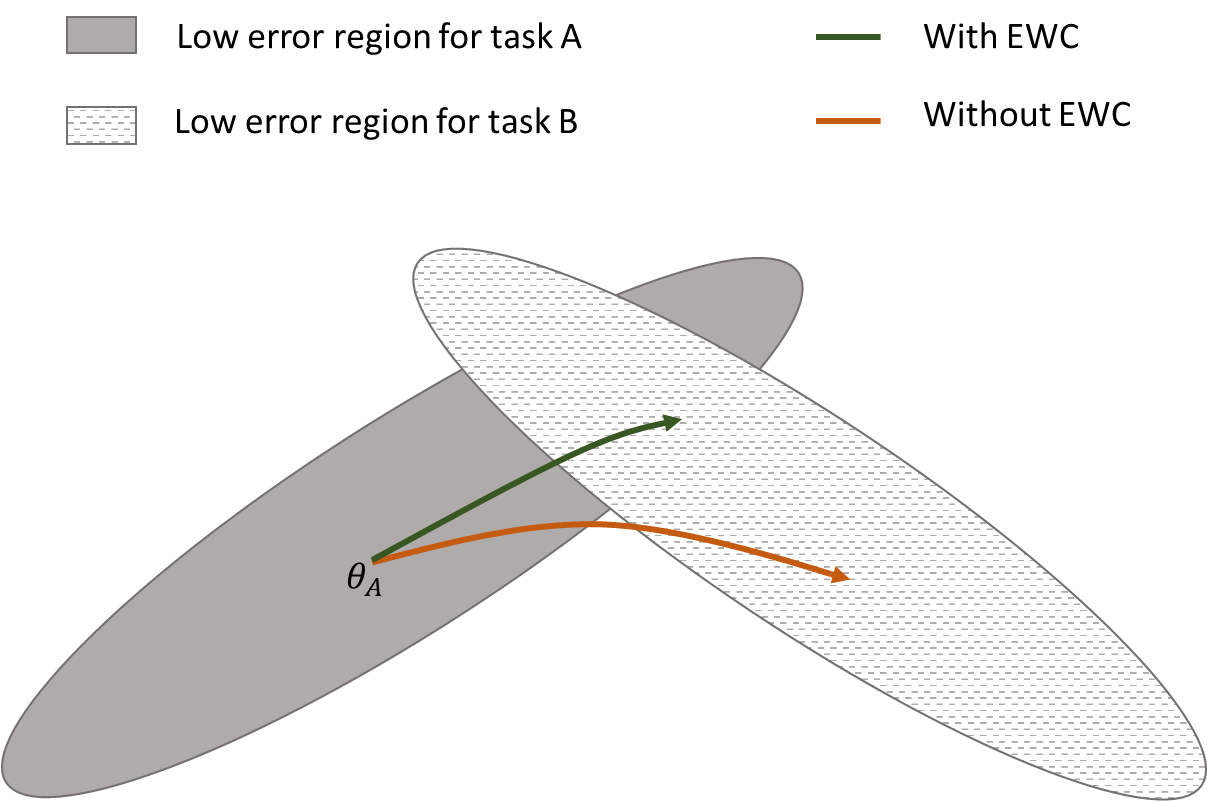}}
  \caption{Training trajectory for task B of a NN algorithm trained for task A with and without EWC}
  \label{ResEWC}
\end{figure}

A schematically representation of a probable training trajectory with and without EWC is given in \cref{ResEWC}.
\subsection{Generation Constraint Violation in NN Loss Function}\label{AP_GenNN}
Generation Constraint violations are included in the NN loss function as follows 
\begin{equation}
    \mathcal{L} (\theta) = \mathcal{L}_T (\theta) + \sum_i \Lambda_g \big (|| \sigma ( G - \overline{G}) ||_2^2 + || \sigma (\underline{G}- {G}) ||_2^2 \big)
\end{equation}
where $ \mathcal{L}_T (\theta)$ is the MAE in the training set, $\Lambda_g$ is the weight assigned to the generation constraint violation in the loss function, and $\sigma$ is the ReLU activation function.  

\bibliographystyle{IEEEtran}
\bibliography{Journal}

\vfill

\end{document}